\documentclass[10pt,twocolumn,letterpaper]{article}

\usepackage[pagenumbers]{cvpr} 

\usepackage{graphicx}
\usepackage{amsmath}
\usepackage{amssymb}
\usepackage{booktabs}

\usepackage{multirow}
\usepackage[normalem]{ulem}

\usepackage[pagebackref,breaklinks,colorlinks]{hyperref}

\usepackage[capitalize]{cleveref}
\crefname{section}{Sec.}{Secs.}
\Crefname{section}{Section}{Sections}
\Crefname{table}{Table}{Tables}
\crefname{table}{Tab.}{Tabs.}

\def\cvprPaperID{11330} 
\def\confName{CVPR}
\def\confYear{2022}

\begin{document}

\title{Improving Visual Grounding with Visual-Linguistic Verification \\and Iterative Reasoning}

\author{Li Yang\textsuperscript{1,2$^*$}, 
Yan Xu\textsuperscript{3$^*$}, 
Chunfeng Yuan\textsuperscript{1$\dagger$}, 
Wei Liu\textsuperscript{1}, 
Bing Li\textsuperscript{1}, 
and Weiming Hu\textsuperscript{1,2,4}\\
\textsuperscript{1}NLPR, Institute of Automation, Chinese Academy of Sciences\\
\textsuperscript{2}School of Artificial Intelligence, University of Chinese Academy of Sciences\\
\textsuperscript{3}The Chinese University of Hong Kong\\
\textsuperscript{4}CAS Center for Excellence in Brain Science and Intelligence Technology\\
{\tt\small\{li.yang,cfyuan,bli,wmhu\}@nlpr.ia.ac.cn, yanxu@link.cuhk.edu.hk, liuwei@ia.ac.cn}}

\maketitle

\renewcommand{\thefootnote}{}
\footnotetext[1]{$*$ denotes equal contribution.} \footnotetext[1]{$\dagger$ denotes the corresponding author. }

\begin{abstract}

Visual grounding is a task to locate the target indicated by a natural language expression. Existing methods extend the generic object detection framework to this problem. They base the visual grounding on the features from 
pre-generated proposals or anchors, and fuse these features with the text embeddings to locate the target mentioned by the text. 
However, modeling the visual features from these predefined locations may fail to fully exploit the visual context and attribute information in the text query, which limits their performance. 
In this paper, we propose a transformer-based framework for accurate visual grounding by establishing text-conditioned discriminative features and performing multi-stage cross-modal reasoning. 
Specifically, we develop a visual-linguistic verification module to focus the visual features on regions relevant to the textual descriptions while suppressing the unrelated areas. A language-guided feature encoder is also devised to aggregate the visual contexts of the target object to improve the object's distinctiveness.  
To retrieve the target from the encoded visual features, we further propose a multi-stage cross-modal decoder to iteratively speculate on the correlations between the image and text for accurate target localization. 
Extensive experiments on five widely used datasets 
validate the efficacy of our proposed components and demonstrate state-of-the-art performance. Our code is public at \url{https://github.com/yangli18/VLTVG}.

\end{abstract}

\section{Introduction}
\label{sec:intro}

\begin{figure}[t]
\centering
\includegraphics[width=0.8\columnwidth]{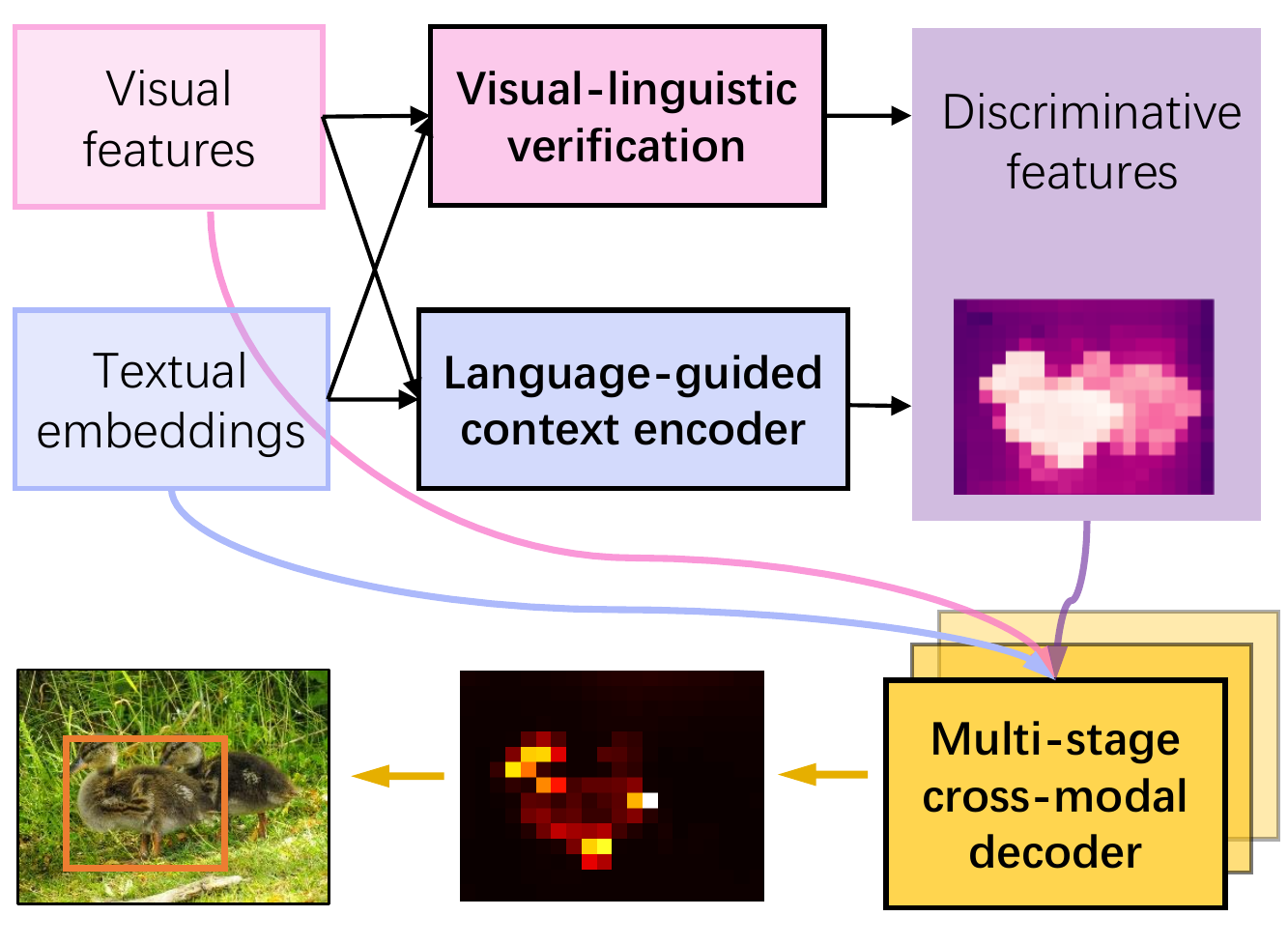}
\caption{Our proposed framework for visual grounding. With the features from the two modalities as input, the visual-linguistic verification module and language-guided context encoder establish discriminative features for the referred object. Then, the multi-stage cross-modal decoder iteratively mulls over all the visual and linguistic features to identify and localize the object. 
}
\label{fig:intro}
\end{figure}

Visual grounding aims to localize the referred object or region in an image by a natural language expression. 
This task has received increasing attention because of its great potential in bridging the gap between linguistic expressions and visual perception. 
The evolution of this technique is also of great importance to other multi-modal reasoning tasks. 
In visual grounding, the referred object is generally specified by one or more pieces of information in the language expression. The information may include object categories, appearance attributes, and visual relation contexts, \etc.
Thus, to avoid ambiguity in reasoning, it is crucial to fully exploit the textual information and model discriminative visual features for visual grounding. 



Existing methods, no matter the two-stage or one-stage ones, treat visual grounding as a ranking problem on the detected candidate regions. 
The two-stage methods~\cite{mao2016generation,nagaraja2016modeling,wang2018learning,yu2016modeling} generally first detect a set of object proposals and then match them with the language query to retrieve the top-ranked proposal. The one-stage approaches~\cite{chen2018real,liao2020real,yang2019fast} directly fuse the text embeddings with image features to generate dense detections, from which to choose the one with the highest confidence score.
As these methods are based on generic object detectors, their inference procedures rely on the predictions from all possible candidate regions, 
which makes the performance limited by the quality of the pre-predicted proposals or the 
configuration of predefined anchor boxes. 
Moreover, they represent the candidate objects with the region features (corresponding to the predicted proposals) or the point features (of the dense anchor boxes) to match or fuse with the text embeddings. Such feature representations may be less flexible for capturing detailed visual concepts or contexts mentioned in linguistic descriptions. 
This inflexibility may increase the difficulties in recognizing the target object. 
Although some methods exploit modular attention~\cite{yu2018mattnet}, graph and tree~\cite{wang2019neighbourhood,yang2019dynamic,yang2020graph,liu2019learning} to better model the relations between vision and language, their processing pipelines are complicated and the performance is still limited by the object proposal inputs. 

Recently, the boom of the transformer in natural language processing~\cite{vaswani2017attention,devlin2018bert} and computer vision~\cite{carion2020end,dosovitskiy2020image} has shown its powerful modeling capability in both the language and vision fields. 
Motivated by that, TransVG~\cite{deng2021transvg} proposes a transformer-based framework for visual grounding. Taking the visual and linguistic feature tokens as inputs, they stack a set of transformer encoder layers to perform cross-modal fusion, and directly predict the target locations. 
Despite its effectiveness, the shared transformer encoder layers are in charge of multiple tasks, 
including encoding the visual-linguistic features, identifying the object instances, and acquiring the final locations, 
which may increases the learning difficulty and could only achieve compromised results. 
It is also less straightforward to retrieve the visual features of the target object with their feature fusion mechanism. 
Thus, we propose to establish a more dedicated framework for accurate visual grounding.



In this work, we propose a transformer-based visual grounding framework that directly retrieves the target object's feature representation for localization.  To this end, as shown in Fig.~\ref{fig:intro}, we first establish the discriminative feature representations by visual-linguistic verification and context aggregation, and then identify the referred object by multi-stage reasoning. Specifically, the visual-linguistic verification module compares the visual features with the semantic concepts from textual embeddings to focus on the regions relevant to the language expression. 
In parallel, the language-guided context encoder gathers the context features to make the visual features of the target object more distinguishable. 
Based on these enhanced features, we propose a multi-stage cross-modal decoder that iteratively compares and mulls over the visual and linguistic information. This enable us to progressively acquire a better representation of the referred object and thereby determine its location more accurately. 


To summarize, our contributions are three-fold: 
(1) To establish the discriminative features for visual grounding, we propose a visual-linguistic verification module to focus the encoded features on the regions related to the language expression. 
A language-guided context encoder is further devised to aggregate important visual contexts for better object identification. 
(2) To retrieve a more accurate feature representation of the referred object, we propose a multi-stage cross-modal decoder, 
which iteratively queries and mulls over visual and linguistic information to reduce the ambiguity during inference.
(3) We benchmark our method on RefCOCO~\cite{yu2016modeling}, RefCOCO+~\cite{yu2016modeling}, RefCOCOg~\cite{mao2016generation}, ReferItGame~\cite{kazemzadeh2014referitgame}, and Flickr30k Entities~\cite{plummer2015flickr30k}. Our method exhibits significant performance improvements over the previous state-of-the-art methods. 
Extensive experiments and ablation studies validate the efficacy of our proposed components. 

\begin{figure*}[t]
	\centering
	\includegraphics[width=0.98\textwidth]{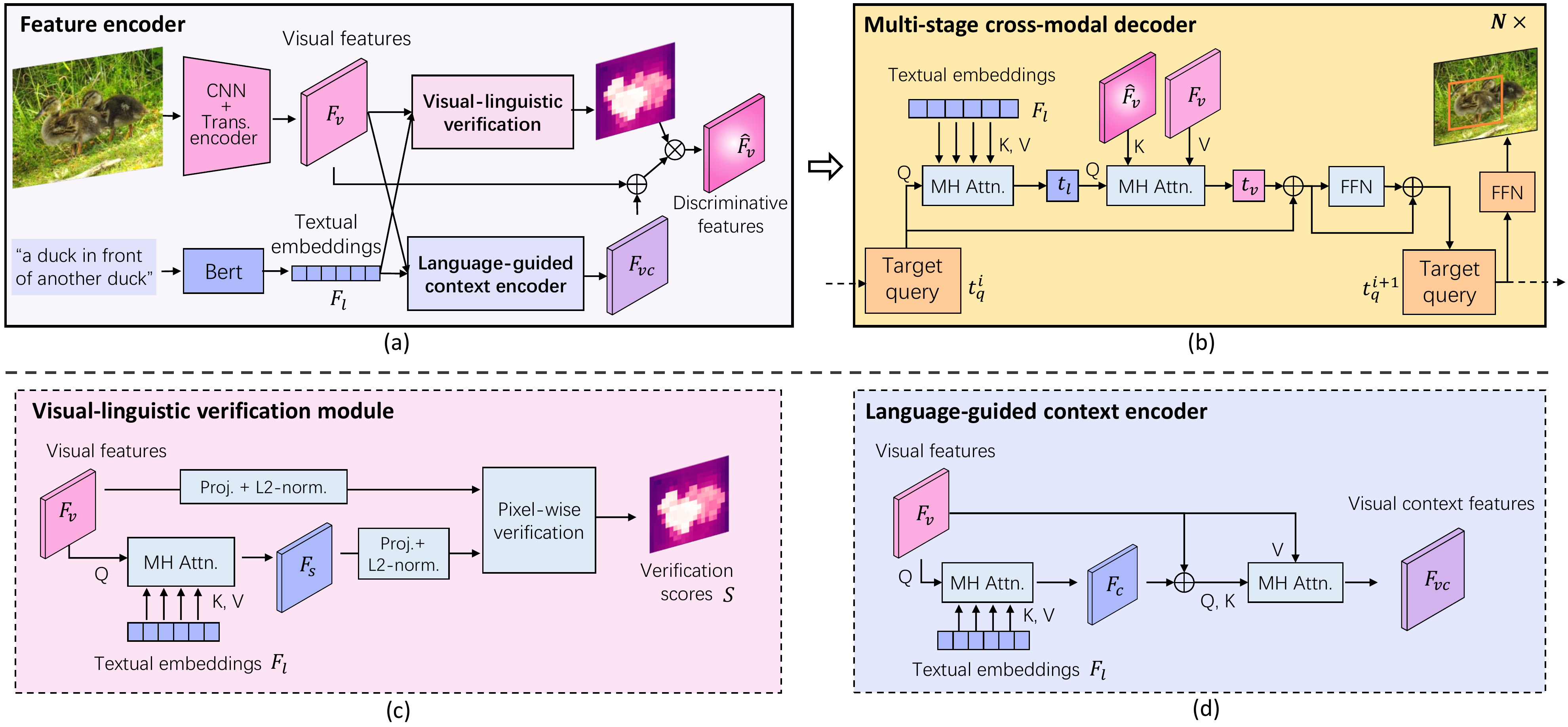}
	\caption{
	The overall network architecture of our proposed method. Given the input image and language expression, the feature encoder (a) first extracts the visual features and textual embeddings, respectively. Then, these features are processed by the visual-linguistic verification module (c) and language-guided context encoder (d) to produce more discriminative features for the referred object. Finally, the multi-stage cross-modal decoder (b) leverages all the visual and textual features to iteratively infer the target location.
	}
	\label{fig:network}
	\vspace{-0.2cm}
\end{figure*}

\section{Related Work}

\subsection{Visual Grounding}
Existing methods generally extend the object detection framework~\cite{ren2016faster,redmon2018yolov3,he2017mask,yang2021pdnet,zhu2020ssn} to address the visual grounding task. 
Two-stage methods~\cite{hong2019learning,hu2017modeling,liu2019learning,wang2018learning,wang2019neighbourhood,yang2019dynamic,yu2018mattnet,zhang2018grounding,zhuang2018parallel} leverage the off-the-shelf detectors to first generate a set of region proposals from the image, and then match them with the language expression to choose the top-ranked proposal. 
However, these methods heavily rely on the performance of the pre-trained detector or proposal generator. If the referred object is not detected 
in the first stage, the ranking and selection process in the second stage will also fail to output the correct detection. 

Recently, one-stage approaches~\cite{chen2018real,liao2020real,yang2020improving,yang2019fast} predict the location of referred object without generating the candidate proposals in advance. For instance, FAOA~\cite{yang2019fast} encodes the language expression into a textual embedding and fuse it into the YOLOv3 detector~\cite{redmon2018yolov3}. The model generates dense object detections with confidence scores, and select the top-ranked one as the prediction for the referred object. 
In order to solve the shortcomings in modeling long language queries, ReSC~\cite{yang2020improving} proposes a iterative sub-query construction framework to reduce the referring ambiguity.
Despite their efficiency, the one-stage methods generally utilize the point feature as the object representation, which may be less flexible to associate with the detailed descriptions in the language expression. 

\subsection{Visual Transformer}

While convolutional neural networks (CNNs) achieved promising results in various vision tasks~\cite{he2016deep,ren2016faster,xu2019depth,xu2021selfvoxelo,xu2022robust,xu2022rnnpose}, the success of transformers in both vision and language fields has attracted great attention of the research community, and the transformer has emerged as a viable alternative to CNNs in many vision tasks, such as image classification~\cite{dosovitskiy2020image}, object detection~\cite{carion2020end,zhu2020deformable}, \etc. 
For example, DETR~\cite{carion2020end} proposes an end-to-end detection framework based on the transformer framework. This work utilizes the attention mechanism of transformer to formulate object detection as a set prediction problem, which obtains outstanding performance. Deformable DETR~\cite{zhu2020deformable} introduces deformable convolution to reduce the computational cost of DETR.  
ViT~\cite{dosovitskiy2020image} applies a pure transformer to image classification and shows excellent performance. 
All these applications in vision tasks demonstrate the powerful and general modeling capabilities of transformer. Motivated by this, we also propose to utilize transformer to develop a more flexible and effective visual grounding framework. 

There are also works exploiting transformer modules or attention mechanisms to solve different types of vision and language tasks~\cite{lee2018stacked,su2021stvgbert,lu2019vilbert,chen2020image}. For example, SCAN~\cite{lee2018stacked} addresses image-text matching by modeling correlations in the proposal-level. STVGBert~\cite{su2021stvgbert} correlates the text embedding with the video frame features for video grounding. In contrast, we focus on visual grounding based on images, and we model the visual-linguistic correlations pixel-wisely to focus feature encoding on the semantically related regions for grounding.


\vspace{-1ex}
\section{Method}
\label{sec:method}

In this section, we present our proposed 
framework for visual grounding. We first introduce the overall network architecture. Then, we elaborate on our proposed visual-linguistic verification module, language-guided context encoder and multi-stage cross-modal decoder. 
Finally, we detail the loss function for training. 

\subsection{The Overall Network}

Unlike the previous ranking-based visual grounding methods, 
we directly retrieve the object feature representation for object localization. 
As shown in Fig.~\ref{fig:network}, given an image and a language expression, we first input them into two separate branches for feature extraction. For the image, we utilize a convolutional neural network (\eg, ResNet-50~\cite{he2016deep}) with a stack of transformer encoder layers to generate a 2D visual feature map $F_v\in\mathbb{R}^{C\times H\times W}$. 
For the input language expression, we encode it as a sequence of textual embeddings $F_l\in\mathbb{R}^{C\times L}$ with BERT~\cite{devlin2018bert} in the other branch. 
Then, based on the features of both modalities, we apply the visual-linguistic verification module and language-guided context encoder to encode discriminative features $\hat{F}_v\in\mathbb{R}^{C\times H\times W}$ for the referred object. 
The visual-linguistic verification module refines the visual features to focus on the regions related to the referring expression, 
while the language-guided context encoder gathers informative visual contexts to facilitate the target-object identification. 
Finally, a multi-stage cross-modal decoder is applied to iteratively 
mull over the encoded visual and textual features to more exactly retrieve the
object representation for target localization. 

\subsection{Visual-Linguistic Verification Module}
\label{sec:v-l-verify}

The input image is encoded first by the convolutional network and then the follow-up transformer encoder layers to be a visual feature map $F_v$. 
This feature map contains the features of object instances in the image, but has no prior knowledge about the referred object by the language expression. 
Retrieving the representation of the referred object without any prior could be distracted by other objects or regions, causing less accurate localization results. 
To address this issue, we propose the visual-linguistic verification module to compute fine-grained correlations between the visual and textual features and focus features on 
regions related to the semantics in the textual description.

As shown in Fig.~\ref{fig:network}-(c), the visual-linguistic verification module is based on the multi-head attention~\cite{vaswani2017attention}. Here, the visual feature map $F_v\in\mathbb{R}^{C\times H\times W}$ serves as the query and the textual embeddings $F_l\in\mathbb{R}^{C\times L}$ acts as the key and value. 
With the multi-head attention, the related semantic features will be gathered (from the textual embeddings) for each visual feature vector of the visual feature map $F_v$. 
The gathered semantic features are organized as a semantic map $F_s\in\mathbb{R}^{C\times H\times W}$ that is spatially aligned with the visual feature map $F_v$. 
Thereafter, we project both the feature maps $F_v$ and $F_s$ to the same semantic space (via linear projection and L2 normalization) obtaining $F'_v$ and $F'_s$, and compute their semantic correlation as the verification score for each spatial location $(x, y)$ as:
\begin{equation}
  S(x,y) = \alpha\cdot \exp{\left(-\frac{\left(1 - F'_v(x,y)^{\mathrm{T}}F'_s(x,y)\right)^2}{2\sigma^2} \right) },
  \label{eq:verify score}
\end{equation}
where $\alpha$ and $\sigma$ are learnable parameters. 
The verification scores model the semantic relevance of each visual feature to the linguistic expression. Thereby, we are able to establish more salient feature map $\hat{F}_v$ for the referred object by modulating the visual features with the verification scores pixel-wisely:
\begin{equation}
  \hat{F}_v = F_v\cdot S
 \label{eq:verify feat}
\ .
\end{equation}
Such modulated visual features $\hat{F}_v$ naturally suppress regions that are irrelevant to the referring expression, making it easier to recognize and locate the referred object in the later phases.

\subsection{Language-guided Context Encoder}

In addition to establishing the semantic correlations between visual and linguistic representations, modeling the visual contexts (\eg, the interaction relations and relative positions) is also crucial for distinguishing the referred target object from other parts. To this end, we propose a language-guided context encoder to gather context features under the guidance of the textual descriptions.

As illustrated in Fig.~\ref{fig:network}-(d), based on the multi-head attention module (left-most one), we input the visual feature map $F_v$ as the query to collected the relevant semantic information 
from the textual embeddings, 
producing the feature map $F_{c}$ as the corresponding linguistic representation for the visual features. 
Based on  $F_{c}$, 
we employ another multi-head attention module to perform language-guided self-attention for the visual features to encode the involved visual contexts.
Specifically, we take the sum of $F_v$ and $F_c$ as the query and key of the multi-head attention module, to compute the feature correlations in both visual and linguistic representations. 
For each attention head in the multi-head attention module, we formulate the attention value from position $i$ to another position $j$ as:
\begin{equation}
\left\{
\begin{aligned}
& Q = W_{Q}^\mathrm{\,T}(F_v + F_c) \\
& K = W_K^\mathrm{\,T}(F_v + F_c) \\
& \mathrm{attn}_{i,j} = \mathrm{softmax}(\frac{Q(i)^\mathrm{T}(K(j) + W_K^\mathrm{T}R(i-j))}{\sqrt{d_k}})
\end{aligned}
\right.,
\end{equation}
where $W_Q$ and $W_K$ are the linear projection weights for the query and key, $d_k$ is the projection channel dimension, and $R(\cdot)$ is the sinusoidal positional encodings~\cite{vaswani2017attention} of relative positions. 
This language-guided self-attention enables the model to learn to gather important context features for the referred object based on the given textual descriptions. 
The multi-head attention module outputs the map of context feature representations as $F_{vc}$. To establish more discriminative features for the target object, we extend the Eq.~\eqref{eq:verify feat} to fuse both the context features $F_{vc}$ and visual-linguistic verification scores $S$ (from Eq.~\eqref{eq:verify score}) with the visual features $F_v$ as: 
\begin{equation}
  \hat{F}_v = (F_v+F_{vc})\cdot S
\ .
\end{equation}
The produced discriminative feature representations are then utilized in the final multi-stage decoder for target identification and localization. 


\subsection{Multi-stage Cross-modal Decoder}
We leverage the established visual feature maps and the textual embeddings for final target localization. 
To reduce the ambiguity in reasoning, 
we propose a multi-stage cross-modal decoder that iteratively mulls over the visual and linguistic information to distinguish the target object from other parts and retrieve related features for object localization. 


We illustrate the architecture of the decoder in Fig.~\ref{fig:network}-(b). The decoder consists of $N$ stages, each of which is constituted by the same network architecture (with unshared weights) 
for iterative cross-modal reasoning. In the first stage, we employ a learnable 
target query $t^1_q \in \mathbb{R}^{C\times1}$ as the initial representation of the target object. 
This target query is fed into the decoder to extract visual features based on the linguistic expression, and iteratively updates its feature representation into $t^i_q$ ($1\leq i\leq N$) in the following stages.  
Fig.~\ref{fig:network}-(b) shows the feature updating process of each decoder stage. 
Specifically, in the $i$-th stage, the target query $t^i_q$ is input (as the query) to the first multi-head attention module 
to collect informative semantic descriptions about the referred object from the textual embeddings, producing the linguistic representation $t_l \in \mathbb{R}^{C\times1}$. 
Then, in the second multi-head attention module, we use this linguistic representation $t_l$ as the query to compute its correlation with the previously established discriminative features $\hat{F}_v$ (from the feature encoder, Fig.~\ref{fig:network}-(a)), 
and then gather the features of interest from the visual feature map $F_v$. 
In this manner, the gathered visual feature $t_v \in \mathbb{R}^{C\times1}$ for the referred object is produced based on the collected semantic descriptions in $t_l$. 
Thereafter, we use this gathered feature $t_v$ to update the target query $t^i_q$ as:
\begin{equation}
\left\{
\begin{aligned}
& t'_q = \mathrm{LN}(t^i_q+t_v) \\
& t^{i+1}_q = \mathrm{LN}(t'_q + \mathrm{FFN}(t'_q)) \\
\end{aligned}
\right. 
\end{equation}
where $\mathrm{LN}(\cdot)$ denotes the layer normalization, and $\mathrm{FFN}(\cdot)$ is a feed-forward network composed of two linear projection layers with ReLU activations. The updated target query $t^{i+1}_q$ is then input to the next decoder stage for iterative cross-modal reasoning and feature representation updating. 

Based on this multi-stage structure, the target query $t^i_q$ of each stage is able to focus on different descriptions in the referring expression, and thus more complete and reliable features of the target object can be gathered. 
The target query $t^i_q$ is progressively refined with the gathered features to form a more accurate representation of the target object. 
Finally, we append a three-layer MLPs with ReLU activation functions to the target query output in each stage to predict the bounding box of the referred object. The predicted bounding boxes from all stages are equally supervised to facilitate the multi-stage decoder training.



\subsection{Training Loss}


Our training procedure is more concise and straightforward than the previous ranking-based methods~\cite{yu2018mattnet,yang2020improving}. 
As our network directly regresses the final bounding box, we avoid the positive/negative sample assignment and can directly compute the regression losses with the ground-truth box. Let $\{\hat{b}^i\}_{i=1}^{N}$ denote the predicted bounding boxes from stages $1$ to $N$, and $b$ denote the ground-truth box. 
We sum up the losses between the predicted boxes and the ground-truth box over all the stages as: 
\begin{equation}
\mathcal{L} = \sum_{i=1}^{N} \lambda_{\mathrm{giou}} \mathcal{L}_{\mathrm{giou}}(b, \hat{b}^i) + \lambda_{\mathrm{L1}} \mathcal{L}_{\mathrm{L1}}(b, \hat{b}^i)
\ ,
\label{eq_loss}
\end{equation}
where $\mathcal{L}_{\mathrm{giou}}(\cdot,\cdot)$ and $\mathcal{L}_{\mathrm{L1}}(\cdot,\cdot)$ are the GIoU loss~\cite{rezatofighi2019generalized} and L1 loss  respectively. The $\lambda_{\mathrm{giou}}$ and $\lambda_{\mathrm{L1}}$ are the hyper-parameters to balance the two losses during training.





\section{Experiments}

\subsection{Datasets}

\noindent\textbf{RefCOCO/ RefCOCO+/ RefCOCOg.}
RefCOCO~\cite{yu2016modeling}, RefCOCO+~\cite{yu2016modeling}, and RefCOCOg~\cite{mao2016generation} are three commonly used benchmarks for visual grounding, with images collected from the MS COCO dataset~\cite{lin2014microsoft}. 
1) RefCOCO~\cite{yu2016modeling} 
has 19,994 images with 142,210 referring expressions for 50,000 referred objects. 
It is split into train, validataion, testA, and testB sets with 120,624, 10,834, 5,657, and 5,095 expressions, respectively.
2) RefCOCO+~\cite{yu2016modeling} provides 19,992 images and 141,564 referring expressions for 49,856 referred objects. It is also split into the train, validation, testA, and testB sets that have 120,191, 10,758, 5,726, and 4,889 referring expressions, respectively. 
3) RefCOCOg~\cite{mao2016generation} has 25,799 images with 95,010 referring expressions for 49,822 object instances. 
The expressions in RefCOCOg are generally longer than those in the other two datasets. There are two splitting conventions for RefCOCOg, which are RefCOCOg-google~\cite{mao2016generation} (val-g) and RefCOCOg-umd~\cite{nagaraja2016modeling} (val-u, test-u). We conduct experiments following the two conventions separately for comprehensive comparisons.

\noindent\textbf{ReferItGame.} ReferItGame~\cite{kazemzadeh2014referitgame} contains 20,000 images from the SAIAPR-12 dataset~\cite{escalante2010segmented}. 
We follow the previous works~\cite{hu2016natural,yang2020improving} to split the dataset into three subsets with 54,127, 5,842 and 60,103 referring expressions for training, validation and testing, respectively. 

\noindent\textbf{Flickr30k Entities.} Flickr30k Entities~\cite{plummer2015flickr30k} has 31,783 images and 427k referred targets. We follow the same split as in the previous works~\cite{plummer2015flickr30k,yang2020improving,deng2021transvg} for training, validation, and testing.

\begin{table*}[t]
\renewcommand{\arraystretch}{1.1168}
\centering
\caption{Comparison of our method with other state-of-the-art methods on RefCOCO~\cite{yu2016modeling}, RefCOCO+~\cite{yu2016modeling}, and RefCOCOg~\cite{mao2016generation}. 
}
\label{tab:sota}
\resizebox{0.7168\textwidth}{!}{%
\begin{tabular}{c|c|ccc|ccc|ccc}
\hline
\multirow{2}{*}{Models}     & \multirow{2}{*}{Backbone} & \multicolumn{3}{c|}{RefCOCO} & \multicolumn{3}{c|}{RefCOCO+}                  & \multicolumn{3}{c}{RefCOCOg}                      \\
    &   & \textit{val} & \textit{testA} & \textit{testB} & \textit{val} & \textit{testA} & \textit{testB} & \textit{val-g} & \textit{val-u} & \textit{test-u} \\ \hline
\textit{Two-stage:}             &                           &              &                &                &              &                &                &                &                &                 \\
CMN\cite{hu2017modeling}        & VGG16                     & -            & 71.03          & 65.77          & -            & 54.32          & 47.76          & 57.47          & -              & -               \\
VC\cite{zhang2018grounding}     & VGG16                     & -            & 73.33          & 67.44          & -            & 58.40          & 5.318           & 62.30          & -              & -               \\
ParalAttn~\cite{zhuang2018parallel}     & VGG16                     & -            & 75.31          & 65.52          & -            & 61.34          & 50.86          & 58.03          & -              & -               \\
MAttNet~\cite{yu2018mattnet}            & ResNet-101                & 76.65        & 81.14          & 69.99          & 65.33        & 71.62          & 56.02          & -              & 66.58          & 67.27           \\
LGRANs~\cite{wang2019neighbourhood}     & VGG16                     & -            & 76.60          & 66.40          & -            & 64.00          & 53.40          & 61.78          & -              & -               \\
DGA~\cite{yang2019dynamic}              & VGG16                     & -            & 78.42          & 65.53          & -            & 69.07          & 51.99          & -              & -              & 63.28           \\
RvG-Tree~\cite{hong2019learning}        & ResNet-101                & 75.06        & 78.61          & 69.85          & 63.51        & 67.45          & 56.66          & -              & 66.95          & 66.51           \\
NMTree~\cite{liu2019learning}           & ResNet-101                & 76.41        & 81.21          & 70.09          & 66.46        & 72.02          & 57.52          & 64.62          & 65.87          & 66.44           \\
Ref-NMS~\cite{chen2021ref}              & ResNet-101    & 80.70     & 84.00     & 76.04     & 68.25     & 73.68       & 59.42     & -          & 70.55    & 70.62           \\ \hline
\textit{One-stage:}         &                           &              &                &                &              &                &                &                &                &                 \\
SSG~\cite{chen2018real}             & DarkNet-53                & -            & 76.51          & 67.50          & -            & 62.14          & 49.27          & 47.47          & 58.80          & -               \\
FAOA~\cite{yang2019fast}            & DarkNet-53                & 72.54        & 74.35          & 68.50          & 56.81        & 60.23          & 49.60          & 56.12          & 61.33          & 60.36           \\
RCCF~\cite{liao2020real}            & DLA-34                    & -            & 81.06          & 71.85          & -            & 70.35          & 56.32          & -              & -              & 65.73           \\
ReSC-Large~\cite{yang2020improving} & DarkNet-53    & 77.63     & 80.45     & 72.30     & 63.59     & 68.36     & 56.81     & 63.12     & 67.30     & 67.20           \\
LBYL-Net~\cite{huang2021look}       & DarkNet-53   & 79.67     & 82.91     & 74.15     & 68.64     & 73.38     & 59.49     & 62.70     & -         & -           \\ \hline
\textit{Transformer-based:}         &               &              &                &                &              &                &                &                &                &                 \\
TransVG~\cite{deng2021transvg}      & ResNet-50     & 80.32     & 82.67     & 78.12     & 63.50     & 68.15     & 55.63     & 66.56     & 67.66     & 67.44           \\
TransVG~\cite{deng2021transvg}      & ResNet-101    & 81.02     & 82.72     & 78.35     & 64.82     & 70.70     & 56.94     & 67.02     & 68.67     & 67.73           \\ \hline
VLTVG (ours)        & ResNet-50   & \textbf{84.53} & \textbf{87.69} & \textbf{79.22}    & \textbf{73.60} & \textbf{78.37} & \textbf{64.53}     & \textbf{72.53}     & \textbf{74.90} & \textbf{73.88}     \\ 
VLTVG (ours)        & ResNet-101  & \textbf{84.77} & \textbf{87.24} & \textbf{80.49}   & \textbf{74.19} & \textbf{78.93} & \textbf{65.17}    & \textbf{72.98} & \textbf{76.04}    & \textbf{74.18}    \\ \hline
\end{tabular}%
}
\end{table*}

\subsection{Implementation Details}

We set the input image size to $640\times640$ and the maximum length of the language expression to $40$. During inference, we resize the input image to make the longer edge equal to $640$ and pad the shorter edge to $640$. 
We append the [CLS] and [SEP] tokens to the beginning and end of the input language expression respectively, before processing the expression with BERT~\cite{devlin2018bert}. 

During training, we use the AdamW optimizer~\cite{loshchilov2018decoupled} to train our proposed model with a batch size of 64. 
We set the initial learning rate of our network to $10^{-4}$, except for the feature extraction branches (\ie the CNN + transformer encoder layers, and BERT), which have an initial learning rate of $10^{-5}$. 
We use ResNet-50 or ResNet-101 as our CNN backbone followed by 6 transformer encoder layers in the visual feature extraction branch, which are initialized with the corresponding weights of DETR model~\cite{carion2020end}. The textual embedding extraction branch is initialized with BERT~\cite{devlin2018bert}. 
We use Xavier init~\cite{glorot2010understanding} to randomly initialize the parameters of the other components in our network. 
We train our model for 30 epochs in all ablation studies, where the learning rate decays by a factor of 10 after 20 epochs. For comparison with state-of-the-art methods, we extend the training epochs to 90, and decay the learning rate by 10 after 60 epochs.
To stabilize the training, we also freeze the weights of the visual and textual feature extraction branches in the first 10 epochs. 
For the learnable parameters in Eq.~\eqref{eq:verify score}, we set $\alpha=1.0$ and $\sigma=0.5$ as the initial values.
For the loss function in Eq.~\eqref{eq_loss}, we set $\lambda_{\mathrm{giou}}=2$ and $\lambda_{\mathrm{L1}}=5$.
We follow the previous works~\cite{deng2021transvg,yang2020improving,liao2020real,yang2019fast} to perform data augmentation during training.

We follow the evaluation metric of the previous works~\cite{yang2020improving,deng2021transvg}. 
Given an image and a language expression, the predicted bounding box is considered correct if its IoU with the ground-truth bounding box is larger than $0.5$.

\subsection{Comparisons with State-of-the-art Methods}

In Table~\ref{tab:sota}, we report our performance in comparison with other state-of-the-art methods on RefCOCO~\cite{yu2016modeling}, RefCOCO+~\cite{yu2016modeling}, and RefCOCOg~\cite{mao2016generation} datasets. 
Our method 
outperforms other methods in all splits of the three datasets. Notably, we achieve absolute improvements of up to $4.45\%$, $5.94\%$ and $5.49\%$ respectively on RefCOCO, RefCOCO+, and RefCOCOg,
compared with the cutting-edge two-stage method Ref-NMS~\cite{chen2021ref}. 
The recent two-stage methods~\cite{yu2018mattnet,chen2021ref} rely on a pre-trained object detector, \eg, Mask R-CNN~\cite{he2017mask} for object proposal generation and feature extraction. 
Since their object detector is not involved in visual grounding training, the visual features from the detector may not be very compatible with the visual grounding task. 
Moreover, the quality of the pre-generated proposals could also become the performance bottleneck of these two-stage paradigms. In contrast, our method encodes the target's feature representation with the guidance from the language expression, which is more flexible and adaptive. 

Our method also shows consistent improvements over the current one-stage methods~\cite{yang2020improving,huang2021look}. 
Notably, we achieve accuracies of $84.77\%$ and $80.49\%$ on the val and testB splits of RefCOCO, 
which bring absolute improvements of $5.10$ and $6.34$ percentage points over the previous best performance of the one-stage framework~\cite{huang2021look}. 
The previous one-stage methods use the fused visual-textual feature at each spatial location to represent a candidate object, which may not adequately capture the important visual attributes or contexts in the language expression. 
Instead, our method models the representation of the target object by iteratively querying the linguistic and visual features, allowing us to identify the referred object in a finer-grained way. 

Comparing with the the most recent work TransVG~\cite{deng2021transvg}, 
our method also achieves appreciable improvements. 
As shown in Table~\ref{tab:sota}, our accuracies exceed  
TransVG  by $2.14\%\!\sim\!4.52\%$  on RefCOCO, $8.23\%\!\sim\!9.37\%$ on RefCOCO+, and $5.96\%\!\sim\!7.37\%$  on RefCOCOg. 

Table~\ref{tab:sota-refit-flickr} also reports the performance of our method on the test sets of the ReferItGame~\cite{kazemzadeh2014referitgame} and Flickr30k Entities~\cite{plummer2015flickr30k}. 
Our method surpasses the previous one-stage and two-stage methods by an appreciable margin. Compared with TransVG~\cite{deng2021transvg}, we achieve relatively better performance with all the different backbones. We notice that our improvements over TransVG on Flickr30k Entities are less than the improvements on the other datasets. This could be because most of the referring expressions in Flickr30k Entities are short noun phrases, which may be not suitable for exhibiting the advantages of our method. 
Nevertheless, 
the experimental results demonstrate the generality and competitiveness of our method in different scenarios.

\begin{table}[t]
\renewcommand{\arraystretch}{1.0}
\centering
\caption{Comparison with the state-of-the-art methods on the test sets of ReferItGame~\cite{kazemzadeh2014referitgame} and Flickr30k Entities~\cite{plummer2015flickr30k}.}
\label{tab:sota-refit-flickr}
\resizebox{0.928\columnwidth}{!}{%
\begin{tabular}{c|c|c|c}
\hline
Models & Backbone & \multicolumn{1}{c|}{\begin{tabular}[c]{@{}c@{}}ReferItGame\\ test\end{tabular}} & \multicolumn{1}{c}{\begin{tabular}[c]{@{}c@{}}Flickr30K\\ test\end{tabular}} \\ \hline
\textit{Two-stage:} &   &   & \\
CMN~\cite{hu2017modeling}       &   VGG16                   &   28.33   &    -   \\
VC~\cite{zhang2018grounding}    &   VGG16                   &   31.13   & -  \\
MAttNet~\cite{yu2018mattnet}    &   ResNet-101              &   29.04   & - \\
Similarity Net~\cite{wang2018learning}    & ResNet-101      &   34.54   & 60.89 \\
CITE~\cite{plummer2018conditional}    &   ResNet-101                      & 35.07           & 61.33 \\
DDPN~\cite{yu2018rethining}     &   ResNet-101          &   63.00       & 73.30 \\ \hline
\textit{One-stage:} &   &   & \\
SSG~\cite{chen2018real}         &   DarkNet-53      &   54.24   &   - \\
ZSGNet~\cite{sadhu2019zero}     &   ResNet-50       &   58.63   &   63.39   \\
FAOA~\cite{yang2019fast}        &   DarkNet-53      &   60.67   &   68.71   \\
RCCF~\cite{liao2020real}        &   DLA-34          &   63.79   &   -   \\
ReSC-Large~\cite{yang2020improving} &   DarkNet-53  &   64.60   &   69.28   \\
LBYL-Net~\cite{huang2021look}   &   DarkNet-53      &   67.47   &   -   \\  \hline
\textit{Transformer-based:} &   &   & \\
TransVG~\cite{deng2021transvg}      &   ResNet-50   &   69.76   &   78.47   \\
TransVG~\cite{deng2021transvg}      &   ResNet-101  &   70.73   &   79.10   \\  \hline
VLTVG (ours)      &   ResNet-50  &   \textbf{71.60}   & \textbf{79.18}   \\
VLTVG (ours)      &   ResNet-101  &  \textbf{71.98}   & \textbf{79.84}   \\  \hline
\end{tabular}%
}
\end{table}

\subsection{Ablation Study}
In this section, we conduct the ablation studies on the RefCOCOg (val-g)~\cite{mao2016generation} dataset, containing longer language expressions, which poses more challenges for comprehension and reasoning.

\vspace{-0.4cm}
\subsubsection{The Component Modules}
In Table~\ref{tab:module-ablation}, we conduct a thorough ablation study on the proposed components to verify their effectiveness. 
The first row of Table~\ref{tab:module-ablation} shows our baseline that performs visual grounding 
with only a single-stage decoder, which achieves an accuracy of $63.64\%$. 
Based on this baseline, we further add more cross-modal decoder stages 
to perform iterative reasoning for target localization. It can be found that the accuracy is notably improved by $2.38$ points, as shown in the second row of Table~\ref{tab:module-ablation}. 
In the third row of Table~\ref{tab:module-ablation}, we further introduce the language-guided context encoder to gather visual contexts for grounding, and find the accuracy is again boosted by $2.42$ points. 
The last row of Table~\ref{tab:module-ablation} shows the performance of our full model after applying the visual-linguistic verification module, 
which further improves the accuracy from $68.44\%$ to $71.62\%$ ($+3.18$ percentage points) and achieves the best performance among these ablation variants.

We also report the number of model parameters and the computational complexity of different variants in Table~\ref{tab:module-ablation}. 
Our proposed components only introduce a total of $8.81$M extra parameters and $0.69$ GFLOPS to the baseline, which only increase the model size and computational complexity by $6.14\%$ and $1.68\%$, respectively. 

\begin{table}[]
\renewcommand{\arraystretch}{1.168}
\centering
\caption{The ablation studies of the proposed components in our network. We evaluate the accuracy of visual grounding, and report the model size and the computational complexity. 
}
\label{tab:module-ablation}
\resizebox{0.938\columnwidth}{!}{%
\begin{tabular}{c|c|c|ccc}
\hline
\multicolumn{1}{c|}{\begin{tabular}[c]{@{}c@{}}Multi-stage\\ Decooder\end{tabular}} & \multicolumn{1}{c|}{\begin{tabular}[c]{@{}l@{}}Context\\ Encoder\end{tabular}} & \multicolumn{1}{c|}{\begin{tabular}[c]{@{}c@{}}V-L\\ Verification\end{tabular}} &  \multicolumn{1}{c}{\#params} & \multicolumn{1}{c}{GFLOPS} & \multicolumn{1}{c}{Acc (\%)} \\ \hline
            &               &               & 143.37M     & 41.10   & 63.64  \\
\checkmark     &               &                & 151.26M     & 41.39 & 66.02    \\
\checkmark     & \checkmark    &                & 151.79M     & 41.67  & 68.44  \\
\checkmark    & \checkmark    & \checkmark      & 152.18M     & 41.79 & \textbf{71.62}   \\ \hline
\end{tabular}%
}
\end{table}

\begin{figure*}[t]
	\centering
	\includegraphics[width=0.96\textwidth]{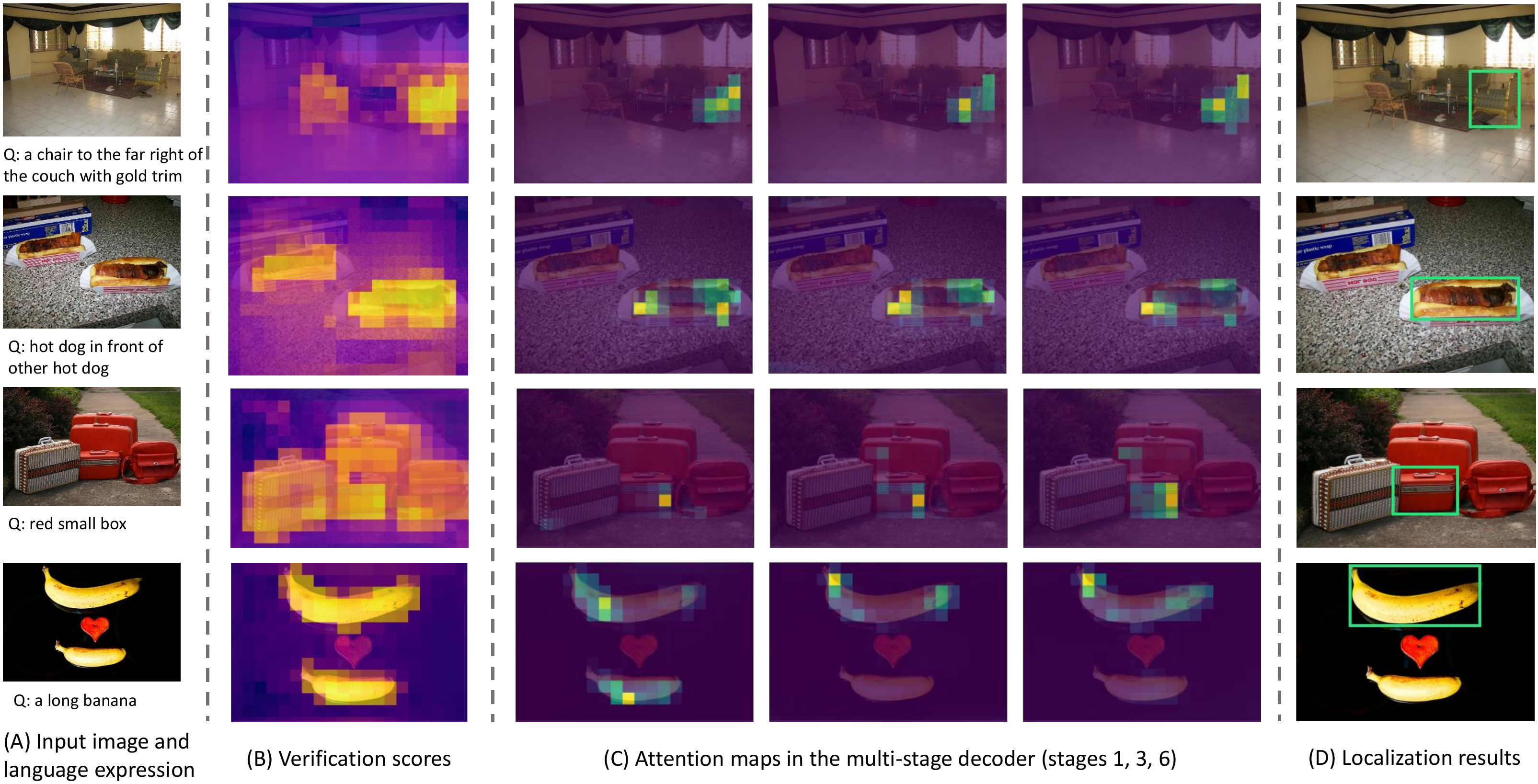}
	\vspace{-0.1cm}
	\caption{
	Visualization of the verification score maps, the multi-stage decoder's attention maps, and the final localization results for various input images and language expressions. 
	}
	\label{fig:vis_result}
	\vspace{-1ex}
\end{figure*}

\subsubsection{The Decoder Stages}


\begin{table}[]
\renewcommand{\arraystretch}{1.16}
\centering
\caption{Comparison of different decoder stages used to perform cross-modal reasoning for visual grounding.}
\label{tab:decoder-stage}
\resizebox{0.86\columnwidth}{!}{%
\begin{tabular}{c|ccc}
\hline
\multicolumn{1}{c|}{The decoder stages ($N$)} & \multicolumn{1}{c}{\#params} & \multicolumn{1}{c}{GFLOPS} & \multicolumn{1}{c}{Acc (\%)}  \\ \hline
$N=1$   & 143.37M   & 41.10     & 63.64         \\
$N=2$   & 144.95M   & 41.15     & 65.05         \\
$N=4$   & 148.10M   & 41.27     & 65.70         \\
$N=6$   & 151.26M   & 41.39     & \textbf{66.02}         \\
$N=8$   & 154.42M   & 41.51     & 65.97         \\ \hline
\end{tabular}%
}
\end{table}

In Table~\ref{tab:decoder-stage}, we evaluate our system when employing different numbers of decoder stages for the object location decoding. 
As shown in Table~\ref{tab:decoder-stage}, the accuracy steadily increases (from $65.22\%$) as more decoder stages are employed until it reaches the saturation point of $67.57\%$ near $N = 6$. 
This reflects the importance of multi-stage reasoning for visual grounding, 
The multi-stage decoder queries the linguistic information and gathers the visual features with multiple rounds, allowing the referred object to be identified and localized more accurately. 
As the accuracy improves little when $N\geq 6$, we employ $6$ decoder stages in our network by default. Besides, our multi-stage decoder only increases the model parameters by $5.5\%$ and the computational cost by $0.71\%$, which is quite efficient.

\begin{table}[]
\renewcommand{\arraystretch}{1.286}
\centering
\caption{Comparison of our method with the transformer-based approach for visual-linguistic feature learning. 
}
\label{tab:vl-feat}
\resizebox{0.928\columnwidth}{!}{%
\begin{tabular}{c|ccc}
\hline
\multicolumn{1}{c|}{{The V-L feature learning}}
                 & \multicolumn{1}{c}{\#params} & \multicolumn{1}{c}{GFLOPS} & \multicolumn{1}{c}{Acc (\%)}                         \\ \hline
None                                    & 151.26M  & 41.39 & 66.02         \\
Trans. encoder layers ($\times 1$)      & 152.69M  & 42.08 & 69.37         \\
Trans. encoder layers ($\times 2$)      & 154.00M  & 42.77 & 69.22         \\
Trans. encoder layers ($\times 3$)      & 155.32M  & 43.46 & 69.15         \\
Trans. encoder layers ($\times 4$)      & 156.63M  & 44.15 & 69.55         \\ \hline
Ours (V-L verification + context)       & 152.18M     & 41.79 & \textbf{71.62}         \\ \hline
\end{tabular}%
}
\end{table}

\subsubsection{Visual-Linguistic Feature Learning}

In our network, we utilize the visual-linguistic verification module and the language-guided context encoder for feature learning of both modalities.
To further verify the necessity and effectiveness, we compare our method with the common transformer-based design for visual-linguistic feature learning in Table~\ref{tab:vl-feat}. 
Specifically, we replace our verification module and context encoder with a stack of common transformer encoder layers. 
We concatenate the visual and textual features and feed them to the stacked transformer encoder layers to perform cross-modal feature fusion. As shown in Table~\ref{tab:vl-feat}, 
using a single transformer encoder layer for feature learning improves the accuracy by $3.35$ points over the baseline. However, when stacking more transformer encoder layers, we observe no significant improvements in accuracy, but a increase in the number of parameters and computational cost. In contrast, our method improves the baseline by $5.60$ percentage points with fewer parameters and computation resources than a single transformer encoder layer. Our method also outperforms the top variant with transformer encoders by $2.07$ percentage points, which demonstrates the efficacy of our design in visual-linguistic features encoding. 

\begin{figure}[t]
\centering
\includegraphics[width=0.886\columnwidth,trim=0 8 0 8,clip]{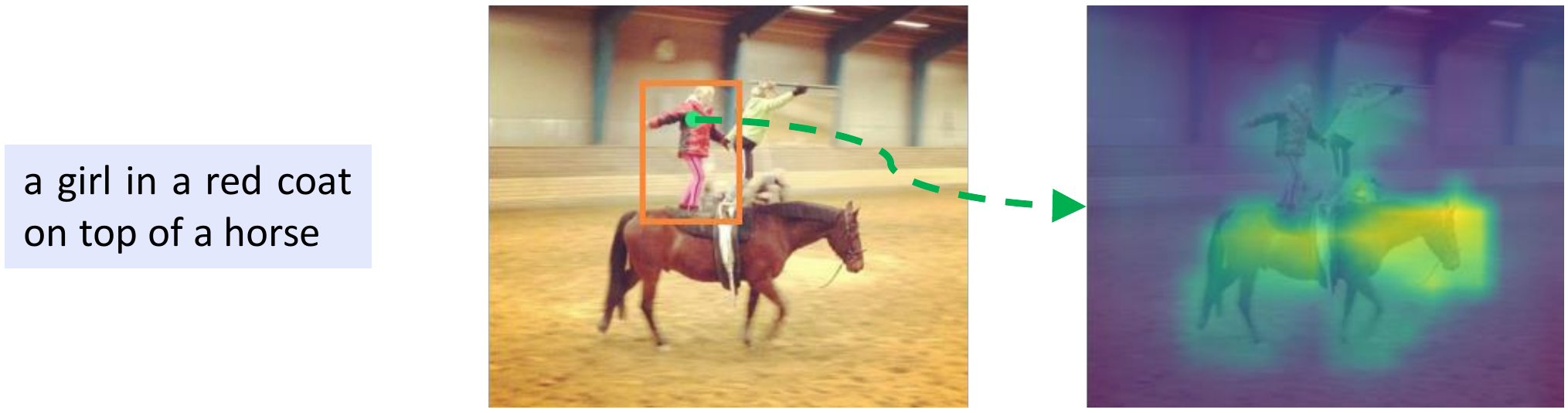}
\vspace{-0.2cm}
\caption{Visualization of the attention map of the language-guided context encoder.}
\label{fig:vis-ctx_enc}
\vspace{-1ex}
\end{figure}

\subsection{Visualization}

In Fig.~\ref{fig:vis_result}, 
we visualize the generated verification scores, the multi-stage decoder's attention maps, and the final localization results for different inputs. It can be observed that, the verification scores are generally higher on the objects or regions related to the descriptions in the language expressions. With Eq.~\eqref{eq:verify feat}, 
our modulated visual features are able to focus on these regions and reduce the distractions. 
Then, the multi-stage decoder mulls over these regions to localize the referred object. For example, in the 4-th row of Fig.~\ref{fig:vis_result}, given the language query ``a long banana", the attention of the first stage is focused on both two bananas. 
After multi-stage reasoning, the attention is successfully drawn to the longer banana mentioned by the text, for localizing the target. 
The discriminative feature representation and multi-stage reasoning enable our method to effectively locate the referred objects for various image and expression inputs. 

In Fig.~\ref{fig:vis-ctx_enc}, given a language expression, we visualize the language-guided context encoder's attention map for a point on the referred girl. Guided by the text, the context encoder focuses on the context around the related horse. The gathered visual contexts are helpful for identifying the girl in the image.


\section{Conclusions and Limitations}
We have presented a transformer-based visual grounding framework, which establishes discriminative features and performs iterative cross-modal reasoning for accurate target localization. 
Our visual-linguistic verification module focuses the visual feature encoding on the regions related to the textual descriptions, while the language-guided context encoder gathers informative visual contexts to improve the distinctiveness of the target. 
Moreover, the multi-stage cross-modal decoder iteratively mulls over the visual and linguistic information 
for localization. 
Extensive experiments on the public datasets exhibit the state-of-the-art performance of our method.

One limitation of this work is that our model is only trained on the visual grounding datasets with limited sizes of corpus. 
Our trained model may not generalize well to more open language expressions. In the future, we plan to extend our method to larger scale grounding datasets for better generalization ability.



\small\noindent\textbf{Acknowledgment} \scriptsize This work is supported by the National Key R\&D Program of China (No. 2018AAA0102802), the Natural Science Foundation of China (Grant No. 61972397, 62036011, 62192782, 61721004), the Key Research Program of Frontier Sciences, CAS, Grant No. QYZDJ-SSW-JSC040, the Featured Innovative Projects of Department of Education of Guangdong Province (Grant No.2019KTSCX175), the China Postdoctoral Science Foundation (Grant No.2021M693402).

{\small
\bibliographystyle{ieee_fullname}
\bibliography{egbib}

\begin{thebibliography}{10}\itemsep=-1pt

\bibitem{carion2020end}
Nicolas Carion, Francisco Massa, Gabriel Synnaeve, Nicolas Usunier, Alexander
  Kirillov, and Sergey Zagoruyko.
\newblock End-to-end object detection with transformers.
\newblock In {\em European Conference on Computer Vision}, pages 213--229.
  Springer, 2020.

\bibitem{chen2021ref}
Long Chen, Wenbo Ma, Jun Xiao, Hanwang Zhang, and Shih-Fu Chang.
\newblock Ref-nms: Breaking proposal bottlenecks in two-stage referring
  expression grounding.
\newblock In {\em Proceedings of the AAAI Conference on Artificial
  Intelligence}, volume~35, pages 1036--1044, 2021.

\bibitem{chen2018real}
Xinpeng Chen, Lin Ma, Jingyuan Chen, Zequn Jie, Wei Liu, and Jiebo Luo.
\newblock Real-time referring expression comprehension by single-stage
  grounding network.
\newblock {\em arXiv preprint arXiv:1812.03426}, 2018.

\bibitem{chen2020image}
Yanbei Chen, Shaogang Gong, and Loris Bazzani.
\newblock Image search with text feedback by visiolinguistic attention
  learning.
\newblock In {\em Proceedings of the IEEE/CVF Conference on Computer Vision and
  Pattern Recognition}, pages 3001--3011, 2020.

\bibitem{deng2021transvg}
Jiajun Deng, Zhengyuan Yang, Tianlang Chen, Wengang Zhou, and Houqiang Li.
\newblock Transvg: End-to-end visual grounding with transformers.
\newblock In {\em Proceedings of the IEEE/CVF International Conference on
  Computer Vision}, pages 1769--1779, 2021.

\bibitem{devlin2018bert}
Jacob Devlin, Ming-Wei Chang, Kenton Lee, and Kristina Toutanova.
\newblock Bert: Pre-training of deep bidirectional transformers for language
  understanding.
\newblock {\em arXiv preprint arXiv:1810.04805}, 2018.

\bibitem{dosovitskiy2020image}
Alexey Dosovitskiy, Lucas Beyer, Alexander Kolesnikov, Dirk Weissenborn,
  Xiaohua Zhai, Thomas Unterthiner, Mostafa Dehghani, Matthias Minderer, Georg
  Heigold, Sylvain Gelly, et~al.
\newblock An image is worth 16x16 words: Transformers for image recognition at
  scale.
\newblock In {\em International Conference on Learning Representations}, 2020.

\bibitem{escalante2010segmented}
Hugo~Jair Escalante, Carlos~A Hern{\'a}ndez, Jesus~A Gonzalez, Aurelio
  L{\'o}pez-L{\'o}pez, Manuel Montes, Eduardo~F Morales, L~Enrique Sucar, Luis
  Villasenor, and Michael Grubinger.
\newblock The segmented and annotated iapr tc-12 benchmark.
\newblock {\em Computer vision and image understanding}, 114(4):419--428, 2010.

\bibitem{glorot2010understanding}
Xavier Glorot and Yoshua Bengio.
\newblock Understanding the difficulty of training deep feedforward neural
  networks.
\newblock In {\em Proceedings of the thirteenth international conference on
  artificial intelligence and statistics}, pages 249--256, 2010.

\bibitem{he2017mask}
Kaiming He, Georgia Gkioxari, Piotr Doll{\'a}r, and Ross Girshick.
\newblock Mask r-cnn.
\newblock In {\em Proceedings of the IEEE international conference on computer
  vision}, pages 2961--2969, 2017.

\bibitem{he2016deep}
Kaiming He, Xiangyu Zhang, Shaoqing Ren, and Jian Sun.
\newblock Deep residual learning for image recognition.
\newblock In {\em Proceedings of the IEEE conference on computer vision and
  pattern recognition}, pages 770--778, 2016.

\bibitem{hong2019learning}
Richang Hong, Daqing Liu, Xiaoyu Mo, Xiangnan He, and Hanwang Zhang.
\newblock Learning to compose and reason with language tree structures for
  visual grounding.
\newblock {\em IEEE transactions on pattern analysis and machine intelligence},
  2019.

\bibitem{hu2017modeling}
Ronghang Hu, Marcus Rohrbach, Jacob Andreas, Trevor Darrell, and Kate Saenko.
\newblock Modeling relationships in referential expressions with compositional
  modular networks.
\newblock In {\em Proceedings of the IEEE Conference on Computer Vision and
  Pattern Recognition}, pages 1115--1124, 2017.

\bibitem{hu2016natural}
Ronghang Hu, Huazhe Xu, Marcus Rohrbach, Jiashi Feng, Kate Saenko, and Trevor
  Darrell.
\newblock Natural language object retrieval.
\newblock In {\em Proceedings of the IEEE conference on computer vision and
  pattern recognition}, pages 4555--4564, 2016.

\bibitem{huang2021look}
Binbin Huang, Dongze Lian, Weixin Luo, and Shenghua Gao.
\newblock Look before you leap: Learning landmark features for one-stage visual
  grounding.
\newblock In {\em Proceedings of the IEEE/CVF Conference on Computer Vision and
  Pattern Recognition}, pages 16888--16897, 2021.

\bibitem{kazemzadeh2014referitgame}
Sahar Kazemzadeh, Vicente Ordonez, Mark Matten, and Tamara Berg.
\newblock Referitgame: Referring to objects in photographs of natural scenes.
\newblock In {\em Proceedings of the 2014 conference on empirical methods in
  natural language processing (EMNLP)}, pages 787--798, 2014.

\bibitem{lee2018stacked}
Kuang-Huei Lee, Xi Chen, Gang Hua, Houdong Hu, and Xiaodong He.
\newblock Stacked cross attention for image-text matching.
\newblock In {\em Proceedings of the European Conference on Computer Vision
  (ECCV)}, pages 201--216, 2018.

\bibitem{liao2020real}
Yue Liao, Si Liu, Guanbin Li, Fei Wang, Yanjie Chen, Chen Qian, and Bo Li.
\newblock A real-time cross-modality correlation filtering method for referring
  expression comprehension.
\newblock In {\em Proceedings of the IEEE/CVF Conference on Computer Vision and
  Pattern Recognition}, pages 10880--10889, 2020.

\bibitem{lin2014microsoft}
Tsung-Yi Lin, Michael Maire, Serge Belongie, James Hays, Pietro Perona, Deva
  Ramanan, Piotr Doll{\'a}r, and C~Lawrence Zitnick.
\newblock Microsoft coco: Common objects in context.
\newblock In {\em European Conference on Computer Vision}, pages 740--755.
  Springer, 2014.

\bibitem{liu2019learning}
Daqing Liu, Hanwang Zhang, Feng Wu, and Zheng-Jun Zha.
\newblock Learning to assemble neural module tree networks for visual
  grounding.
\newblock In {\em Proceedings of the IEEE/CVF International Conference on
  Computer Vision}, pages 4673--4682, 2019.

\bibitem{loshchilov2018decoupled}
Ilya Loshchilov and Frank Hutter.
\newblock Decoupled weight decay regularization.
\newblock In {\em International Conference on Learning Representations}, 2018.

\bibitem{lu2019vilbert}
Jiasen Lu, Dhruv Batra, Devi Parikh, and Stefan Lee.
\newblock Vilbert: Pretraining task-agnostic visiolinguistic representations
  for vision-and-language tasks.
\newblock {\em Advances in neural information processing systems}, 32, 2019.

\bibitem{mao2016generation}
Junhua Mao, Jonathan Huang, Alexander Toshev, Oana Camburu, Alan~L Yuille, and
  Kevin Murphy.
\newblock Generation and comprehension of unambiguous object descriptions.
\newblock In {\em Proceedings of the IEEE conference on computer vision and
  pattern recognition}, pages 11--20, 2016.

\bibitem{nagaraja2016modeling}
Varun~K Nagaraja, Vlad~I Morariu, and Larry~S Davis.
\newblock Modeling context between objects for referring expression
  understanding.
\newblock In {\em European Conference on Computer Vision}, pages 792--807.
  Springer, 2016.

\bibitem{plummer2018conditional}
Bryan~A Plummer, Paige Kordas, M~Hadi Kiapour, Shuai Zheng, Robinson Piramuthu,
  and Svetlana Lazebnik.
\newblock Conditional image-text embedding networks.
\newblock In {\em Proceedings of the European Conference on Computer Vision
  (ECCV)}, pages 249--264, 2018.

\bibitem{plummer2015flickr30k}
Bryan~A Plummer, Liwei Wang, Chris~M Cervantes, Juan~C Caicedo, Julia
  Hockenmaier, and Svetlana Lazebnik.
\newblock Flickr30k entities: Collecting region-to-phrase correspondences for
  richer image-to-sentence models.
\newblock In {\em Proceedings of the IEEE international conference on computer
  vision}, pages 2641--2649, 2015.

\bibitem{redmon2018yolov3}
Joseph Redmon and Ali Farhadi.
\newblock Yolov3: An incremental improvement.
\newblock {\em arXiv preprint arXiv:1804.02767}, 2018.

\bibitem{ren2016faster}
Shaoqing Ren, Kaiming He, Ross Girshick, and Jian Sun.
\newblock Faster r-cnn: Towards real-time object detection with region proposal
  networks.
\newblock {\em IEEE Transactions on Pattern Analysis and Machine Intelligence},
  39(6):1137--1149, 2016.

\bibitem{rezatofighi2019generalized}
Hamid Rezatofighi, Nathan Tsoi, JunYoung Gwak, Amir Sadeghian, Ian Reid, and
  Silvio Savarese.
\newblock Generalized intersection over union: A metric and a loss for bounding
  box regression.
\newblock In {\em Proceedings of the IEEE/CVF Conference on Computer Vision and
  Pattern Recognition}, pages 658--666, 2019.

\bibitem{sadhu2019zero}
Arka Sadhu, Kan Chen, and Ram Nevatia.
\newblock Zero-shot grounding of objects from natural language queries.
\newblock In {\em Proceedings of the IEEE/CVF International Conference on
  Computer Vision}, pages 4694--4703, 2019.

\bibitem{su2021stvgbert}
Rui Su, Qian Yu, and Dong Xu.
\newblock Stvgbert: A visual-linguistic transformer based framework for
  spatio-temporal video grounding.
\newblock In {\em Proceedings of the IEEE/CVF International Conference on
  Computer Vision}, pages 1533--1542, 2021.

\bibitem{vaswani2017attention}
Ashish Vaswani, Noam Shazeer, Niki Parmar, Jakob Uszkoreit, Llion Jones,
  Aidan~N Gomez, {\L}ukasz Kaiser, and Illia Polosukhin.
\newblock Attention is all you need.
\newblock In {\em Advances in neural information processing systems}, pages
  5998--6008, 2017.

\bibitem{wang2018learning}
Liwei Wang, Yin Li, Jing Huang, and Svetlana Lazebnik.
\newblock Learning two-branch neural networks for image-text matching tasks.
\newblock {\em IEEE Transactions on Pattern Analysis and Machine Intelligence},
  41(2):394--407, 2018.

\bibitem{wang2019neighbourhood}
Peng Wang, Qi Wu, Jiewei Cao, Chunhua Shen, Lianli Gao, and Anton van~den
  Hengel.
\newblock Neighbourhood watch: Referring expression comprehension via
  language-guided graph attention networks.
\newblock In {\em Proceedings of the IEEE/CVF Conference on Computer Vision and
  Pattern Recognition}, pages 1960--1968, 2019.

\bibitem{xu2021selfvoxelo}
Yan Xu, Zhaoyang Huang, Kwan-Yee Lin, Xinge Zhu, Jianping Shi, Hujun Bao,
  Guofeng Zhang, and Hongsheng Li.
\newblock Selfvoxelo: Self-supervised lidar odometry with voxel-based deep
  neural networks.
\newblock In {\em Conference on Robot Learning}, pages 115--125. PMLR, 2021.

\bibitem{xu2022robust}
Yan Xu, Junyi Lin, Jianping Shi, Guofeng Zhang, Xiaogang Wang, and Hongsheng
  Li.
\newblock Robust self-supervised lidar odometry via representative structure
  discovery and 3d inherent error modeling.
\newblock {\em IEEE Robotics and Automation Letters}, 2022.

\bibitem{xu2022rnnpose}
Yan Xu, Junyi Lin, Guofeng Zhang, Xiaogang Wang, and Hongsheng Li.
\newblock Rnnpose: Recurrent 6-dof object pose refinement with robust
  correspondence field estimation and pose optimization.
\newblock {\em arXiv preprint arXiv:2203.12870}, 2022.

\bibitem{xu2019depth}
Yan Xu, Xinge Zhu, Jianping Shi, Guofeng Zhang, Hujun Bao, and Hongsheng Li.
\newblock Depth completion from sparse lidar data with depth-normal
  constraints.
\newblock In {\em Proceedings of the IEEE/CVF International Conference on
  Computer Vision}, pages 2811--2820, 2019.

\bibitem{yang2021pdnet}
Li Yang, Yan Xu, Shaoru Wang, Chunfeng Yuan, Ziqi Zhang, Bing Li, and Weiming
  Hu.
\newblock Pdnet: Towards better one-stage object detection with prediction
  decoupling.
\newblock {\em arXiv preprint arXiv:2104.13876}, 2021.

\bibitem{yang2019dynamic}
Sibei Yang, Guanbin Li, and Yizhou Yu.
\newblock Dynamic graph attention for referring expression comprehension.
\newblock In {\em Proceedings of the IEEE/CVF International Conference on
  Computer Vision}, pages 4644--4653, 2019.

\bibitem{yang2020graph}
Sibei Yang, Guanbin Li, and Yizhou Yu.
\newblock Graph-structured referring expression reasoning in the wild.
\newblock In {\em Proceedings of the IEEE/CVF Conference on Computer Vision and
  Pattern Recognition}, pages 9952--9961, 2020.

\bibitem{yang2020improving}
Zhengyuan Yang, Tianlang Chen, Liwei Wang, and Jiebo Luo.
\newblock Improving one-stage visual grounding by recursive sub-query
  construction.
\newblock In {\em European Conference on Computer Vision}, pages 387--404.
  Springer, 2020.

\bibitem{yang2019fast}
Zhengyuan Yang, Boqing Gong, Liwei Wang, Wenbing Huang, Dong Yu, and Jiebo Luo.
\newblock A fast and accurate one-stage approach to visual grounding.
\newblock In {\em Proceedings of the IEEE/CVF International Conference on
  Computer Vision}, pages 4683--4693, 2019.

\bibitem{yu2018mattnet}
Licheng Yu, Zhe Lin, Xiaohui Shen, Jimei Yang, Xin Lu, Mohit Bansal, and
  Tamara~L Berg.
\newblock Mattnet: Modular attention network for referring expression
  comprehension.
\newblock In {\em Proceedings of the IEEE Conference on Computer Vision and
  Pattern Recognition}, pages 1307--1315, 2018.

\bibitem{yu2016modeling}
Licheng Yu, Patrick Poirson, Shan Yang, Alexander~C Berg, and Tamara~L Berg.
\newblock Modeling context in referring expressions.
\newblock In {\em European Conference on Computer Vision}, pages 69--85.
  Springer, 2016.

\bibitem{yu2018rethining}
Zhou Yu, Jun Yu, Chenchao Xiang, Zhou Zhao, Qi Tian, and Dacheng Tao.
\newblock Rethinking diversified and discriminative proposal generation for
  visual grounding.
\newblock {\em International Joint Conference on Artificial Intelligence
  (IJCAI)}, 2018.

\bibitem{zhang2018grounding}
Hanwang Zhang, Yulei Niu, and Shih-Fu Chang.
\newblock Grounding referring expressions in images by variational context.
\newblock In {\em Proceedings of the IEEE Conference on Computer Vision and
  Pattern Recognition}, pages 4158--4166, 2018.

\bibitem{zhu2020ssn}
Xinge Zhu, Yuexin Ma, Tai Wang, Yan Xu, Jianping Shi, and Dahua Lin.
\newblock Ssn: Shape signature networks for multi-class object detection from
  point clouds.
\newblock In {\em European Conference on Computer Vision}, pages 581--597.
  Springer, 2020.

\bibitem{zhu2020deformable}
Xizhou Zhu, Weijie Su, Lewei Lu, Bin Li, Xiaogang Wang, and Jifeng Dai.
\newblock Deformable detr: Deformable transformers for end-to-end object
  detection.
\newblock In {\em International Conference on Learning Representations}, 2020.

\bibitem{zhuang2018parallel}
Bohan Zhuang, Qi Wu, Chunhua Shen, Ian Reid, and Anton Van Den~Hengel.
\newblock Parallel attention: A unified framework for visual object discovery
  through dialogs and queries.
\newblock In {\em Proceedings of the IEEE Conference on Computer Vision and
  Pattern Recognition}, pages 4252--4261, 2018.

\end{thebibliography}
}

\end{document}